\documentclass[times, twoside, watermark]{zHenriquesLab-StyleBioRxiv}
\usepackage{blindtext}

\leadauthor{Lee}

\begin{document}

\title{Development of collective behavior in newborn artificial agents}
\shorttitle{Development of collective behavior in newborn artificial agents}

\author[1,3]{Donsuk Lee}
\author[1,4]{Samantha M. W. Wood}
\author[1-4]{Justin N. Wood}

\affil[1]{Informatics Department, Indiana University, United States}
\affil[2]{Center for the Integrated Study of Animal Behavior, Indiana University, United States}
\affil[3]{Cognitive Science Program, Indiana University, United States}
\affil[4]{Department of Neuroscience, Indiana University, United States}

\maketitle

\begin{abstract}
Collective behavior is widespread across the animal kingdom. To date, however, the developmental and mechanistic foundations of collective behavior have not been formally established. What learning mechanisms drive the development of collective behavior in newborn animals? Here, we used deep reinforcement learning and curiosity-driven learning—two learning mechanisms deeply rooted in psychological and neuroscientific research—to build “newborn” artificial agents that develop collective behavior. Like newborn animals, our agents learn collective behavior from raw sensory inputs in naturalistic environments. Our agents also learn collective behavior without external rewards, using only intrinsic motivation (curiosity) to drive learning. Specifically, when we raise our artificial agents in natural visual environments with groupmates, the agents spontaneously develop ego-motion, object recognition, and a preference for groupmates, rapidly learning all of the core skills required for collective behavior. This work bridges the divide between high-dimensional sensory inputs and collective action, resulting in a “pixels-to-actions” model of collective animal behavior. More generally, we show that two generic learning mechanisms—deep reinforcement learning and curiosity-driven learning—are sufficient to learn collective behavior from unsupervised natural experience.
\end {abstract}

\begin{keywords}
collective behavior, deep reinforcement learning, curiosity-driven learning, pixels-to-actions, artificial intelligence
\end{keywords}

\begin{corrauthor}
donslee\at iu.edu or woodjn\at indiana.edu
\end{corrauthor}

\section{Introduction}
Collective behavior is widespread across nature. Individuals spontaneously organize into cohesive groups, including schools of fish \cite{aoki_simulation_1982,huth_simulation_1992}, flocks of birds \cite{ballerini_interaction_2008, pearce_role_2014}, and human crowds \cite{silverberg_collective_2013}. From a computational perspective, this is an impressive feat. To engage in collective behavior, newborn animals must solve several challenging computational tasks. First, animals must learn ego-motion (knowledge of one’s location and direction in space), so that they can navigate and unite with their groupmates. Second, animals must learn to detect and recognize groupmates across new viewing situations—an ability known as “invariant” object recognition. Third, animals must develop a \textit{preference} to group, spontaneously reducing the distance between themselves and their groupmates. Remarkably, newborn animals develop all of these abilities rapidly, in the absence of explicit rewards or supervision \cite{wood_newborn_2013}. What learning mechanisms support the rapid development of collective behavior? More generally, how do animals learn to group?

To address these questions, this paper brings together ideas and techniques from two fields that are traditionally separate: (1) the study of collective animal behavior and (2) the study of artificial intelligence in scalable, task-performing embodied agents. We first describe “rule-based” models of collective animal behavior, in which agents are endowed with hard-coded interaction rules. We argue that while rule-based models have been successful for understanding the high-level principles and dynamics of collective behavior, these models are not well suited for understanding the learning mechanisms that underlie collective behavior. To characterize these learning mechanisms, we use \textit{autonomous artificial agents} developed in artificial intelligence. While these artificial agents were not originally designed to be models of collective behavior, we show that these agents spontaneously develop collective behavior, akin to real animals. Specifically, when these artificial agents are raised together in groups, the agents spontaneously develop ego-motion, object recognition, and grouping preferences, solving the core tasks required for collective behavior. This result shows that the development of collective behavior does not require hardcoded interaction rules. Rather, collective behavior can emerge from two core learning mechanisms—deep reinforcement learning and curiosity-driven learning—that are deeply rooted in psychological and neuroscientific research on animal behavior. 

By using tools from artificial intelligence (which were, in turn, inspired by neuroscience), we have isolated a set of biologically-inspired learning mechanisms that are sufficient to generate collective behavior in autonomous agents. We conclude that artificial agents can serve as valuable models of collective behavior in the biological sciences, allowing researchers to formalize the learning mechanisms that generate widespread social behaviors.

\subsection{Rule-Based Models}
To understand the nature and dynamics of collective behavior, researchers typically use rule-based models, where individuals are modeled as featureless points that change their behavior according to a fixed set of interaction rules \cite{couzin_collective_2009,vicsek_novel_1995}. Rule-based models have provided many valuable insights into collective behavior. For example, collective behavior can emerge from repeated local interactions \cite{aoki_simulation_1982,couzin_self-organization_2003,reynolds_flocks_1987,vicsek_novel_1995}. Local interactions allow collective behavior to be distributed and decentralized, making groups inherently robust to perturbation. Rule-based models can also explain how biologically plausible local interactions can account for many group structures found in nature \cite{couzin_collective_2002}.  

According to Tinbergen’s “Four Questions” Framework \cite{tinbergen_aims_1963}, an integrative understanding of a behavior (e.g., collective behavior) requires addressing four complementary questions: (1) What function does the behavior serve? (2) What is the phylogeny of the behavior? (3) What mechanisms support the behavior? (4) How does the behavior develop over the lifetime of the individual? The first two questions (function and phylogeny) address \textit{why} species evolved a particular set of structures and adaptations. The last two questions (mechanisms and development) address \textit{how} individual organisms learn and perform the behavior. Rule-based models are well suited for addressing Tinbergen’s \textit{why} questions because they provide a high-level understanding of the adaptations that emerge from simple interaction rules. However, rule-based models are not well suited for understanding the mechanisms and development of collective behavior, for the reasons described below.

First, rule-based models of collective behavior make simplifying assumptions about the nature of visual perception. Rule-based models generally operate over high-level features (e.g., position and orientation of neighbors) that are not directly available from sensory signals. In contrast, during everyday perception, brains must convert high-dimensional sensory inputs ($10^6$ optic nerve fibers) into a manageably small number of perceptually relevant features that can drive adaptive behavior. To accurately explain how animals \textit{learn} to group, models should solve the same problem as animals and learn from high-dimensional sensory inputs. 

Second, since rule-based models typically use simple, featureless environments, these models do not explain one of the central challenges of collective behavior: detecting and recognizing groupmates in complex, natural environments. Object detection and recognition require complex computational systems that gradually transform streams of sensory data into object-centric representations that can guide adaptive behavior \cite{dicarlo_how_2012, yamins_using_2016}. Ideally, a mechanistic model of collective behavior should be capable of solving the same problem as the brain, parsing groupmates from natural backgrounds and recognizing those groupmates across identify-preserving image transformations (e.g., illumination changes). 

Third, rule-based models of collective behavior typically lack both learning mechanisms and motivational systems to drive learning. Rather, agents follow hardcoded interaction rules and do not learn from their experiences. Real animals, in contrast, depend heavily on learning to develop core visual-motor skills \cite{held_movement-produced_1963}, choosing their own inputs during development to optimize learning \cite{gottlieb_towards_2018, smith_developing_2018}. In sum, while rule-based models of collective behavior are simple and interpretable, these models overlook the underlying complexity of visual perception, do not explain how animals learn to detect and recognize groupmates in naturalistic environments, and do not formalize the learning mechanisms that drive collective behavior. 

Recently, researchers have begun to explore the role of vision and motivational systems in collective behavior. For example, some rule-based models have explored the role of vision in the collective movements of fish \cite{collignon_stochastic_2016, rosenthal_revealing_2015, strandburg-peshkin_visual_2013}, birds \cite{pearce_role_2014}, humans \cite{moussaid_how_2011}, and artificial systems \cite{bastien_model_2020, f_schilling_learning_2019, lavergne_group_2019}. Other models have used agent-based learning systems, in which agents are endowed with an internal mechanism for deciding how to respond to sensory input, and rules for modifying those responses based on past experience \cite{ried_modelling_2019}. Some features of collective motion (coalignment and cohesion) can also emerge from intrinsic motivation, where agents move to maximize the number of visual states they expect to be able to access in the future \cite{charlesworth_intrinsically_2019}. However, none of these models learned from high-dimensional sensory inputs in realistic three-dimensional environments, nor were the models required to detect and recognize groupmates across identity-preserving image transformations: requirements for collective behavior in the real world. 

We suggest that, in order to understand the \textit{development} and \textit{mechanisms} of collective behavior, biologists must turn to a different type of model: scalable “pixels-to-actions” models developed in the field of artificial intelligence. Pixels-to-actions models complement (not replace) rule-based models, providing a unique perspective for exploring how core learning mechanisms interact with complex, naturalistic environments to produce collective behavior in autonomous agents. As a result, pixels-to-actions models allow biologists to tackle Tinbergen’s how (mechanisms and development) questions about collective behavior.

\subsection{Pixels-to-Actions Models}
Our approach involves building “newborn” pixels-to-actions agents that spontaneously develop collective behavior. We use the term “newborn” to refer to a specific class of artificial agents that have four characteristics akin to newborn animals: (a) the artificial agents are embodied in complex (naturalistic) environments, (b) the artificial agents learn from raw sensory inputs, (c) the artificial agents have the capacity to autonomously learn from the environment without direct external supervision, and (d) in the newborn state, the artificial agents have not yet been exposed to a visual environment. Thus, like newborn animals, these artificial agents must solve the entire task, learning from high-dimensional visual inputs (i.e., raw pixels) and performing actions in realistic, three-dimensional (3D) environments. By constructing such pixels-to-actions agents, we can determine which learning mechanisms are sufficient for developing collective behavior in autonomous agents: an essential step toward understanding the developmental and mechanistic foundations of collective action \cite{kriegeskorte_cognitive_2018, newell_you_1973}. As a starting point, we built agents that develop the most general characteristic of collective behavior: group formation and cohesion \cite{ballerini_interaction_2008, lavergne_group_2019}. 

To construct our agents, we focused on two biologically-inspired mechanisms: deep reinforcement learning and curiosity-driven learning. Deep reinforcement learning provides a computational framework for learning adaptive behaviors from high-dimensional sensory inputs. In reinforcement learning, agents maximize their long-term rewards by performing actions in response to their environment and internal state. To succeed in environments approaching real-world complexity, agents must learn abstract and generalizable features to represent the environment. To this end, deep reinforcement learning combines reinforcement learning with deep neural networks in order to transform raw sensory inputs into efficient representations that can support adaptive behavior. Artificial agents trained through deep reinforcement learning can develop human- and animal-like abilities. For example, artificial agents can learn to play simple and complex video games (e.g., Atari: \citenum{mnih_human-level_2015}; Quake III: \citenum{jaderberg_human-level_2019}; StarCraft II: \citenum{vinyals_grandmaster_2019}), develop advanced motor behaviors (e.g., walking, running, and jumping: \citenum{heess_emergence_2017}), and learn to navigate 3D environments \cite{banino_vector-based_2018}.

The second mechanism—curiosity-driven learning—can drive the development of complex behaviors through self-supervised learning \cite{haber_learning_2018, schmidhuber_formal_2010}. Curiosity promotes learning by motivating individuals to seek out informative experiences. By seeking out less predictable (and more informative) experiences, artificial agents can gradually expand their knowledge of the world, continuously acquiring useful training data for improving perception and cognition \cite{oudeyer_how_2016}. A previous study found that a preference for informative states can produce collective behavior (group formation \& coalignment) in groups of agents navigating simple (2D) spaces \cite{charlesworth_intrinsically_2019}. Several studies have also suggested that collective behavior promotes efficient and robust acquisition of environmental information (e.g., predatory threats or food sources; \cite{lemasson_collective_2009, rahmani_flocking_2020, sumpter_information_2008}. We speculate that information-seeking may be a general motivating force underlying group formation in many animal species.

We explored whether deep reinforcement learning and curiosity-driven learning are sufficient to produce collective behavior in complex (3D) visual environments. To test this hypothesis, we leveraged recent advances in artificial intelligence to construct pixels-to-actions agents. Our approach involved embodying deep neural networks in virtual agents, and raising those agents in realistic virtual environments. Like newborn animals, our agents learned to perform actions from raw sensory inputs. Because the agents were active participants in the environment, their learning experiences were self-organized: an agent’s actions determined what its learning inputs would be. Learning in our agents was also not guided by external rewards: our agents were intrinsically motivated to seek informative experiences. Specifically, during development, the agents built an internal model of the dynamics of the environment by learning to predict the consequences of their own actions. Their prediction error was used to reinforce behaviors that led to informative sensory states (those which produced high prediction errors given their current knowledge state). We refer to the combination of these two learning mechanisms as \textit{curiosity-driven deep reinforcement learning}.

\subsection{The Present Experiments}

\begin{figure*}
\centering
\includegraphics[width=\linewidth]{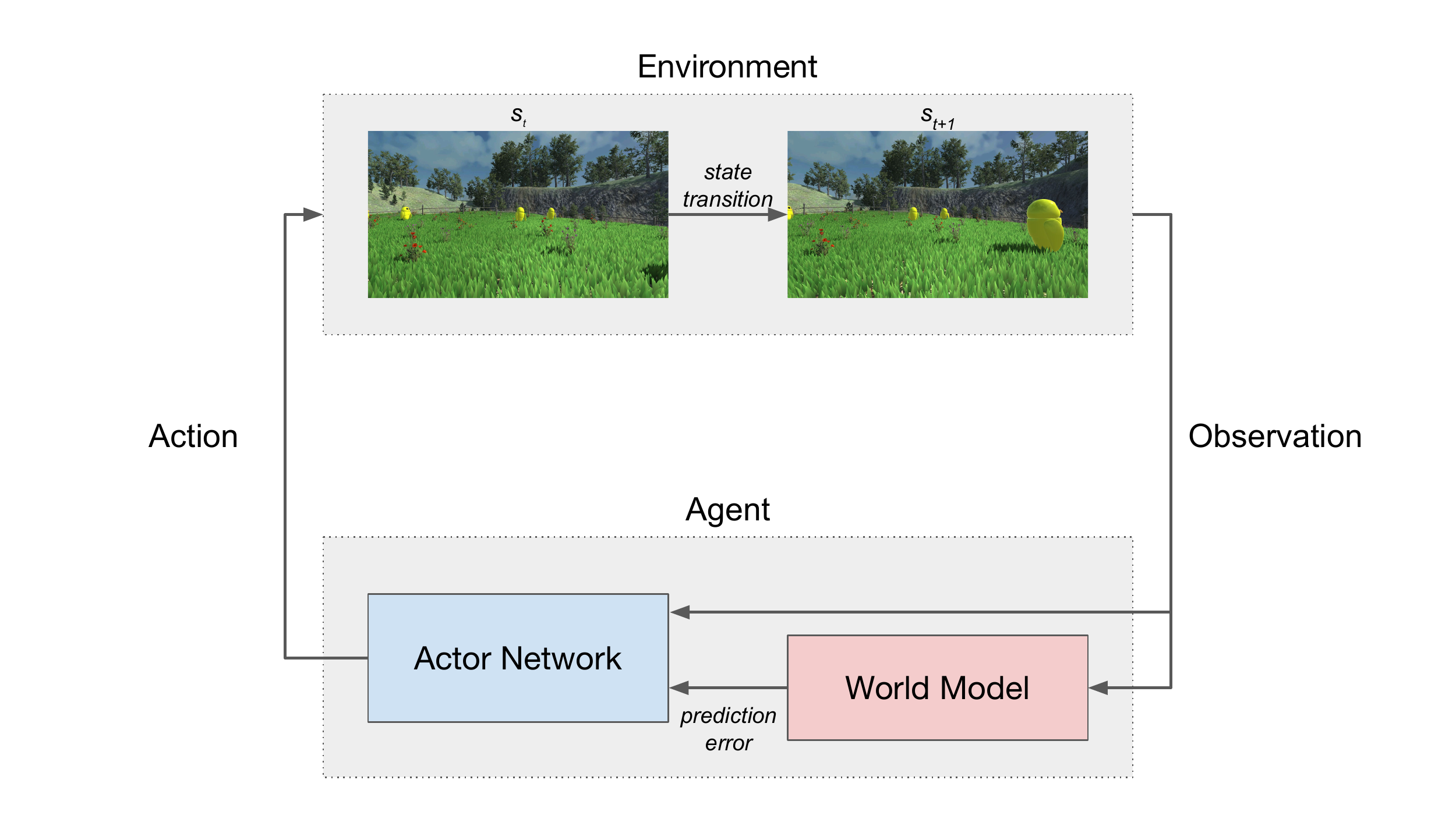}
\caption{
\textbf{Pixels-to-actions model.} Our agents received visual observations and performed actions in virtual environments. No external rewards were provided to the agents. The artificial brains contained two components: (1) a world model and (2) an actor network. The world model learned to make predictions about future observations based on the current observation and action. The prediction error was provided to the actor network as an intrinsic reward, instantiating a form of curiosity-driven learning. The actor network learned a policy (a function mapping an observation to an action) that maximized cumulative intrinsic reward using deep reinforcement learning.
}
\label{fig:model}
\end{figure*}

To explore whether curiosity-driven deep reinforcement learning is sufficient to produce collective behavior, we created autonomous artificial agents and raised those agents together in groups (akin to newborn animals, who are typically raised with conspecifics). As a starting point, we trained and tested our artificial agents in simple environments (Experiments 1-2), and then extended the approach to a more complex, realistic environment, which was more similar to environments experienced by animals in nature (Experiments 3-4). In Experiment 5, we then examined whether our agents could develop grouping behavior in conditions that closely matched those used to study newborn animals in laboratory studies. 

Our artificial brains contained two core modules: (i) a world model that learned to predict environmental dynamics and generate intrinsic reward (curiosity) signals based on prediction error, and (ii) an actor network that learned to select actions based on raw visual inputs and intrinsic rewards through deep reinforcement learning (\cref{fig:model}). The world model roughly corresponds to the “ventral visual-hippocampal stream” in animals, which processes visual input through a series of hierarchically-organized areas and guides information-seeking behavior through intrinsic rewards relayed through the entorhinal-hippocampal complex \cite{santos-pata_entorhinal_2021}. Conversely, the actor network roughly corresponds to the “dorsal visual stream” in animals, which transforms visual input into an egocentric coordinate system for visually-guided action. Both modules were constructed using deep neural networks. There is extensive evidence that deep neural networks can serve as powerful computational models of the ventral and dorsal visual streams. For example, deep neural networks can accurately predict the response properties of neurons in the ventral visual stream \cite{yamins_using_2016, zhuang_unsupervised_2021}. Researchers have also used features from deep reinforcement learning agents (akin to those used here) to predict neural activation patterns in the dorsal visual stream \cite{cross_using_2021}. In the light of such empirical evidence, we used deep neural networks in our model of collective behavior.

At each time step of the simulation, the agent received visual input from the environment and performed an action. While interacting with the environment, the agent continuously adjusted its internal model of environmental dynamics (the world model) to predict how sensory inputs would change in response to its own actions. In particular, the agent learned to predict how its actions would impact its visual observations. We provided the prediction error to the reinforcement learning algorithm as the intrinsic reward \cite{burda_large-scale_2019, pathak_curiosity-driven_2017}. The agent learned which actions to perform based on maximizing the cumulative intrinsic reward. Because the groupmates were the least predictable parts of the environment, we hypothesized that the intrinsic curiosity reward would motivate the agents to be interested in one another, leading to the spontaneous development of collective behavior.

To preview our results, we show that when learning is subject to a critical period (akin to the learning of social preferences in animals), these artificial agents spontaneously develop collective behavior. We also show that these agents can learn collective behavior in complex, realistic, and visually rich environments. Finally, we demonstrate that collective behavior can emerge in a variety of artificial brains, including brains with small, medium, and large neural networks. These results provide a developmental model of collective behavior: an important step towards understanding how autonomous animals spontaneously learn to engage in collective action.

\section{Experiment 1}

\begin{figure*}
\centering
\includegraphics[width=\linewidth]{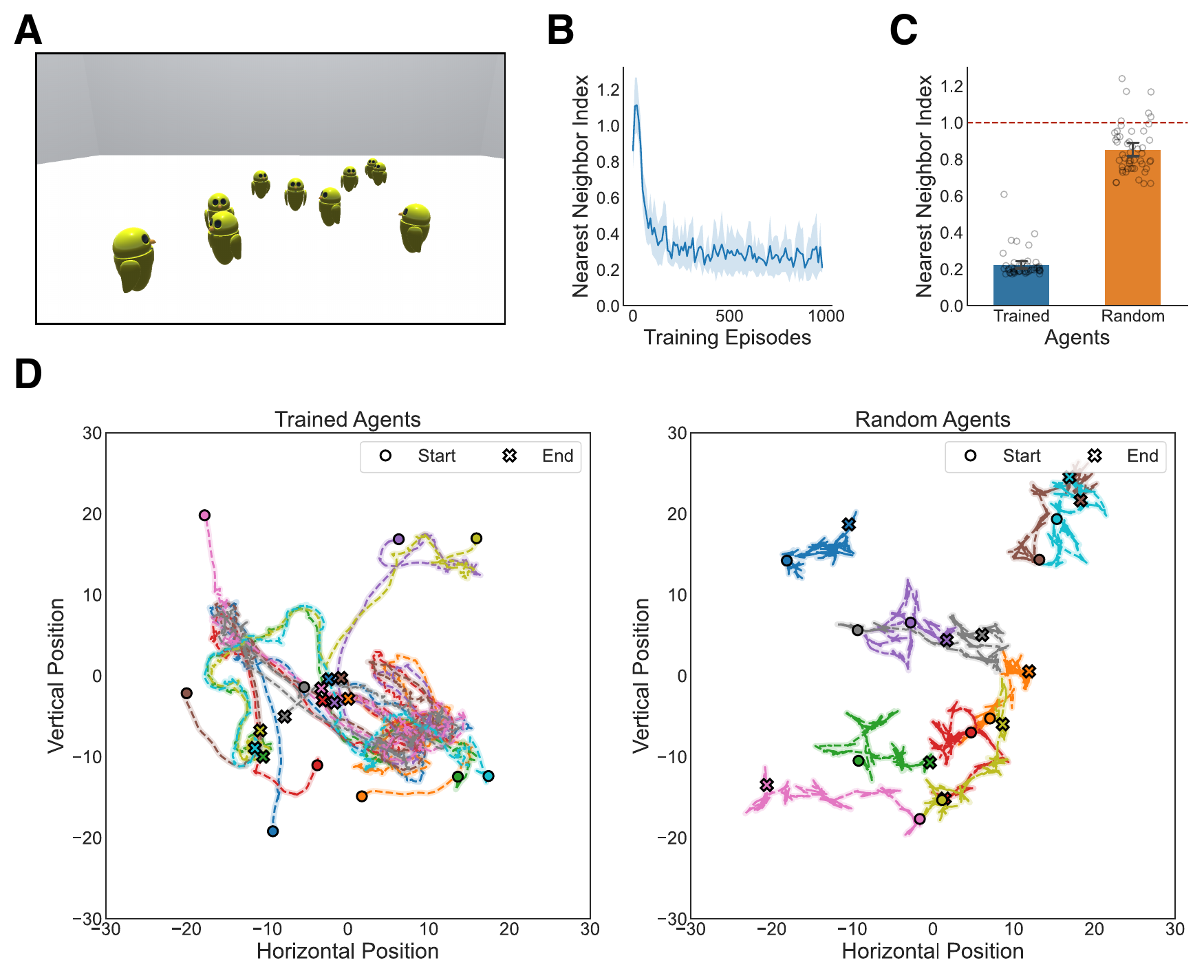}
\caption{\textbf{Experiment 1.} \textbf{(A)} The artificial agents. Each agent received raw visual inputs through an invisible camera mounted on its head. The agents were trained together in the same environment (a white cubic chamber). \textbf{(B)} The nearest neighbor index (NNI) across the Training Phase. Initially, the agents were more likely to explore (leading to an increase in the NNI). As the training progressed, the agents began to seek out other agents and exhibited grouping behavior (leading to a decrease in the NNI). \textbf{(C)} Comparison between the NNIs of trained agents and random agents. The red dashed line represents the expected NNI of randomly-distributed agents. \textbf{(D)} Trajectories of trained agents (left) and random agents (right) during the Test Phase. The agents started at random positions in the environment (denoted by circle markers). As the trial progressed, the trained agents exhibited grouping behavior as they sought out other agents. By the end of the test trials (denoted by “X” markers), the trained agents had grouped together, whereas the random agents remained dispersed.}
\label{fig:exp1}
\end{figure*}

We first explored whether grouping behavior can emerge from the combination of deep reinforcement learning and curiosity-driven learning. To develop grouping behavior, the agents needed to learn to move (ego-motion), detect and recognize groupmates (object recognition), and approach those groupmates (preference for grouping).

\subsection{Virtual Environments}
We created the virtual environments using the Unity game engine. The environments and code can be accessed at \url{https://buildingamind.github.io/grouping/}. For Experiment 1, we created an environment called “Simple World.” Simple World was a white cubic chamber with 60×60×60m (L×W×H) walls and a light source at the center of the chamber ceiling.

\subsection{Agents}
We created the artificial agents by embodying artificial brains in virtual animal bodies. The animal bodies were modelled after newborn birds. Each agent had a yellow body, which fit into a cylinder measuring 3.5 unit (height) and 1.2 unit length (radius). Each agent received visual input (96×96 pixel resolution images) through an invisible forward-facing camera attached to its head. The agents could move forward or backwards, rotate left or right, or remain stationary. The actions were represented as a pair of discrete variables: translation along the longitudinal axis and rotation around the vertical axis. Each action incurred a metabolic cost to incentivize the agents to move smoothly and efficiently. The action costs were chosen to reflect the fact that backward locomotion elicits a greater metabolic demand than forward locomotion \cite{flynn_comparison_1994}. The costs for forward motion, backward motion, rotation, and idling were -0.001, -0.01, -0.0005, and 0.0 respectively.

The architecture of the world model was adapted from \cite{pathak_curiosity-driven_2017}. The world model learned to predict the environmental dynamics using self-supervision and produced intrinsic rewards based on its prediction errors. The world model consisted of three neural network components: a feature encoder $\phi$, an inverse dynamics model $g$, and a forward dynamics model $f$, parametrized by $\theta_\phi$, $\theta_g$ and $\theta_f$, respectively. Given a transition $(s_t,a_t,s_{t+1})$, where $s_t$ and $s_{t+1}$ are visual observations at consecutive time steps and $a_t$ is an action taken at time $t$ , the visual observations were encoded into representations $x_t=\phi(s_t;\theta_\phi)$ and $x_{t+1}=\phi(s_{t+1};\theta_\phi)$ by the feature encoder. The inverse model learned to predict actions $\hat{a}_t=g(x_t,x_{t+1};\theta_g)$ from the features of two consecutive observations. The forward model learned to predict the features of the observation at time $t+1$, $\hat{x}_(t+1)=f(x_t,a_t;\theta_f)$, from the features $x_t$ and action $a_t$ at time $t$. The parameters $\theta_\phi$, $\theta_g$ and $\theta_f$ were jointly optimized to minimize the weighted average of inverse and forward dynamics prediction errors:
$$
(1-\alpha)L_i\left[ g(x_t,x_{t+1};\theta_g),a_t \right] + \alpha L_f\left[ f(x_t,a_t;\theta_f),x_{t+1} \right],
$$
where $\alpha$ is a weighting factor, $L_i$ is the negative log-likelihood of the true action $a_t$ according to the inverse model’s prediction, and $L_f$ is the squared error between the forward model’s prediction and the features $x_{t+1}$. The value of $\alpha$ was set to 0.2 for all experiments in this paper. The forward model’s prediction error was multiplied by a scaling factor and provided to the actor network as the intrinsic reward.

The feature encoder of the world model was a convolutional neural network (CNN) followed by a fully-connected layer with 128 units. The inverse dynamics model consisted of a hidden layer with 256 hidden units and two output heads to predict translation and rotation actions. The forward dynamics model was a 2-layer MLP with 256 hidden units and 128 output units.

The actor network $\pi$ received visual observation $s_t$ at time $t$ and sampled an action $a_t$ from a stochastic policy $\pi(s_t;\theta_p)$, parameterized by $\theta_p$. In our case with a discrete action space, $\pi(s_t;\theta_p)$ was a categorical probability distribution across possible actions. Given a transition $(s_t,a_t,s_{t+1})$ between $t$ and $t+1$, the reward function was defined as:
$$
R_t (s_t,a_t,s_{t+1})=R_m(a_t) + R_c(s_t,a_t,s_{t+1}),
$$
where $R_m$ is a metabolic cost of action $a_t$ and $R_c$ is the intrinsic reward (i.e., curiosity) output from the world model  taking the transition as input. The actor network was optimized to maximize the expected sum of rewards $\mathbb{E}[\sum_{t} \gamma^t R_t]$ with discount factor $\gamma$. Thus, the actor network learned to seek sensory states that were the least predictable for the world model.

The actor network processed visual inputs through a CNN feature encoder followed by two fully-connected layers with 128 hidden units. Then, the outputs from the last hidden layer were passed into two separate output layers to produce a vector of probabilities for each translation action (forward, backward, no translation) and each rotation action (clockwise, counterclockwise, no rotation). 

In Experiment 1, we used a CNN with 2 convolution layers for the feature encoder of the world model, and a CNN with 15 convolution layers (with skip connections) for the feature encoder of the actor network. In Experiment 4, we explored whether other neural network architectures also produce collective behavior. We created our agents using Unity ML-Agents Toolkit version 0.10.1 \cite{juliani_unity_2020}. 

\subsection{Training Phase}
We trained and tested 10 artificial agents simultaneously within the large cubic room (\cref{fig:exp1}A). All of the trained agents had the same brain architecture. However, each agent’s brain started with a different random initialization of the connection weights within the architecture, and each agent’s connection weights were shaped by its own particular experiences during the Training Phase. The agents received no external rewards from the environment. The agents’ actions were motivated entirely by intrinsic rewards: curiosity and a metabolic cost associated with movements.

At the beginning of the Training Phase, the agents were spawned at random positions and orientations within their environment. Each training episode lasted for 1,000 time steps. The agents were reset at random positions and orientations at the beginning of each episode. We trained the agents for 1,000 episodes. The agents were trained to optimize the sum of their intrinsic curiosity reward and metabolic cost using Proximal Policy Optimization (PPO; \citenum{schulman_proximal_2017}). We scaled the prediction error of the world model by a factor of 0.1 for the curiosity reward. We used the following hyperparameters in all of our experiments: γ (discount rate) = 0.99, λ (Generalized Advantage Estimate regularization) = 0.95, β (entropy regularization) = 0.001, batch size = 256, buffer size = 2560, learning rate = 0.001. The learning rate decayed linearly, reaching 0 at the end of training. 

\subsection{Test Phase}
After the Training Phase, the model parameters were fixed for the Test Phase (i.e., the agents did not receive any rewards during the Test Phase, and learning was discontinued). Restricting learning to the Training Phase is common in machine learning and also mimics the sensitive period of learning in animals. Sensitive periods are developmental stages when animals are more sensitive to environmental stimuli, such as groupmates. In many species, the tendency to form strong attachments to groupmates is subject to an early sensitive period \cite{horn_pathways_2004}. If our autonomous agents learned grouping behavior during the Training Phase, then they should have formed cohesive groups during the Test Phase, reducing the distance between themselves and the other agents. 

We tested our agents for 50 episodes. Each test episode consisted of 2,000 time steps. At every time step, we recorded the position of each agent in X,Y coordinates. We measured the agents' grouping behavior using the Nearest Neighbor Index (NNI; \citenum{clark_distance_1954}). NNI measures the spatial dispersion of a group of individuals. When agents start at random locations and move randomly, the NNI will be approximately 1; when agents cluster together, the NNI will approach 0; and when agents disperse, the NNI will be larger than 1. To compute the NNI, we first calculated the distance between each agent and the nearest neighbor at each time step. Next, we calculated the average of the nearest neighbor distances across agents for each episode. Then, we obtained the NNI for each episode by dividing the average nearest neighbor distance by the nearest neighbor distance expected for a random distribution \cite{clark_distance_1954}.

\subsection{Results and discussion}
The results are shown in Figure 2. The NNI was small (M = 0.22, SD = 0.07; compared to an NNI of 1.0 for a random spatial distribution of agents), indicating that our agents successfully developed ego-motion and grouping behavior. To confirm that the agents showed more pronounced grouping behavior than a group of randomly moving agents, we also tested a group of random, untrained agents. At each time step, the random agents chose an action uniformly at random from their action space, regardless of the visual inputs. The trained agents were significantly more likely to show grouping behavior than the random agents (mean difference = 0.63, SD = 0.11; independent samples \textit{t}-test, \textit{t}(98) = 29.30, $P < 10^{-42}$, Cohen’s \textit{d} = 5.92).

These results indicate that curiosity-driven deep reinforcement learning can generate ego-motion and grouping behavior in artificial agents. However, since the groupmates looked the same across the Training Phase and Test Phase, Experiment 1 does not reveal whether the agents developed invariant recognition of groupmates. In natural visual environments, the appearance of animals can change radically due to changes in viewpoint, illumination, position, etc. To group successfully, animals must therefore build invariant representations of groupmates that tolerate natural image variation. In Experiment 2, we examined whether our artificial agents learned to recognize groupmates across natural image changes.

\section{Experiment 2}
We tested whether our artificial agents build invariant representations of groupmates that tolerate natural image variation (illumination changes). Specifically, in the Test Phase, we systematically changed the illumination angle and the illumination intensity to explore whether the visual representations constructed during the Training Phase would generalize across novel illumination change. 

\begin{figure}[t]
\centering
\includegraphics[width=\linewidth]{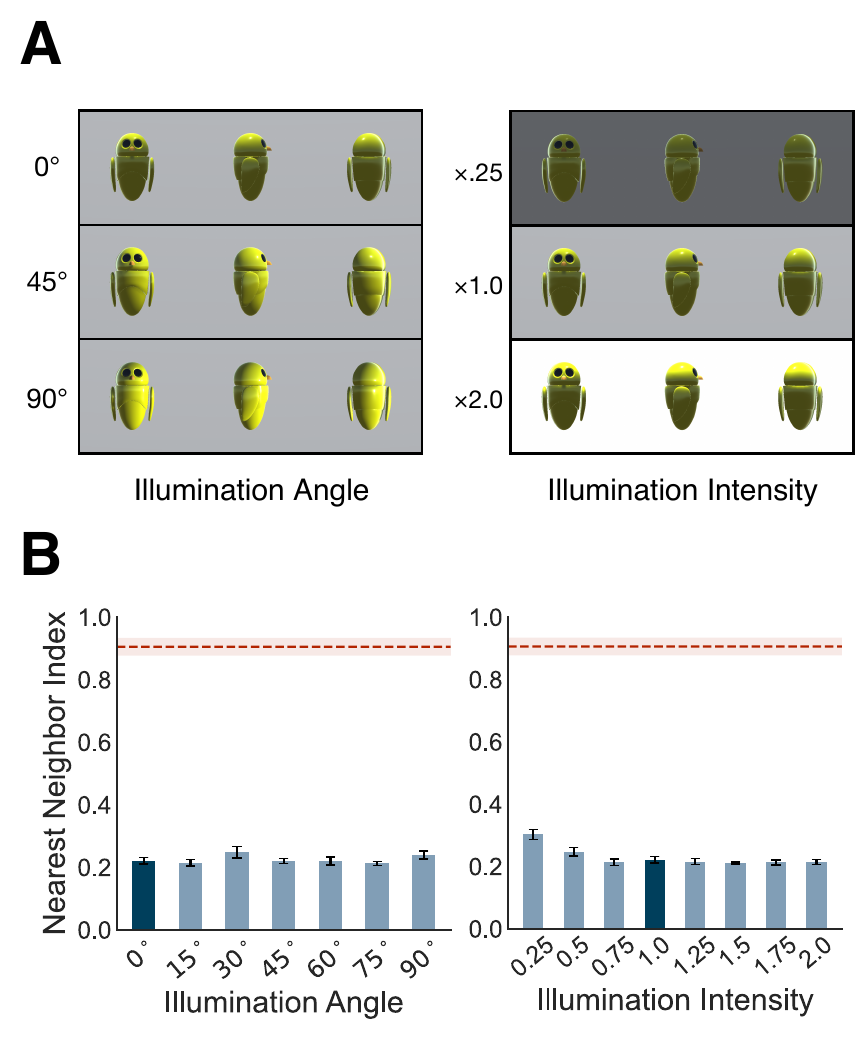}
\caption{\textbf{Experiment 2.} \textbf{(A)} Images of the artificial agents under different illumination conditions (illumination angles, \textit{left}; illumination intensity, \textit{right}). Varying the illumination conditions produced large, novel, and complex changes in the agents’ appearance. All of the agents were trained in a single illumination condition (illumination angle = 0°, illumination intensity = 1.0), and then tested in familiar and novel illumination conditions. \textbf{(B)} Nearest neighbor index (NNI) of the agents across different lighting conditions in the Test Phase. Lower NNIs indicate that the agents were closer together. The agents successfully exhibited grouping behavior under a range of novel illumination conditions. The red lines and associated ribbons show the average and standard error (respectively) of random agents. The dark blue bars represent the illumination conditions from the Training Phase, and the light blue bars represent the novel illumination conditions. }
\label{fig:exp2}
\end{figure}

\subsection{Methods}
The methods were identical to those used in Experiment 1 with one crucial difference: during the Test Phase, we systematically varied the lighting conditions in the virtual chamber. We changed either the illumination angle (\cref{fig:exp2}A, left; +0, +15°, +30°, +45°, +60°, +75°, or +90°) or the illumination intensity (\cref{fig:exp2}A, right; ×0.25, ×0.5, ×0.75, ×1.0, ×1.25, ×1.5, ×1.75, or ×2.0). In the Test Phase, the agents therefore needed to recognize their groupmates across novel illumination conditions. We then computed the average NNI for the trained agents and compared the trained agents to the random agents.

\subsection{Results and Discussion}
The trained agents showed significantly more pronounced grouping behavior than the random agents in all of the illumination angle conditions (\cref{fig:exp2}B, left; independent samples t-tests, all $P$s < $10^{-39}$) and illumination intensity conditions (\cref{fig:exp2}B, right; independent samples t-tests, all $P$s < $10^{-36}$). The agents developed robust grouping behavior, despite the fact that the agents looked significantly different during the Training Phase versus Test Phase. Thus, during training, our agents built visual representations that generalized across novel illumination conditions. These results indicate that our agents developed a form of invariant recognition using purely self-supervised learning rules, and that these representations supported grouping behavior even when the agents experienced visual conditions that they had never encountered before.  

Together, Experiments 1-2 indicate that two learning mechanisms—deep reinforcement learning and curiosity-driven learning—are sufficient for generating robust grouping behavior in simple environments. In the real world, however, animals must learn to recognize groupmates in complex, natural environments \cite{rahmani_flocking_2020}. Natural environments introduce a number of challenges for biological and artificial agents. For instance, natural environments contain a variety of conspicuous features (e.g., rocks, trees) that could distract agents and impair their grouping behavior. Natural environments also make it more difficult for agents to detect and recognize groupmates because the visual features of those agents must be parsed—or segmented—from the visual features in the background. Can our artificial agents overcome these challenges and learn grouping behavior in naturalistic environments?

\section{Experiment 3}

\begin{figure*}
\centering
\includegraphics[width=\linewidth]{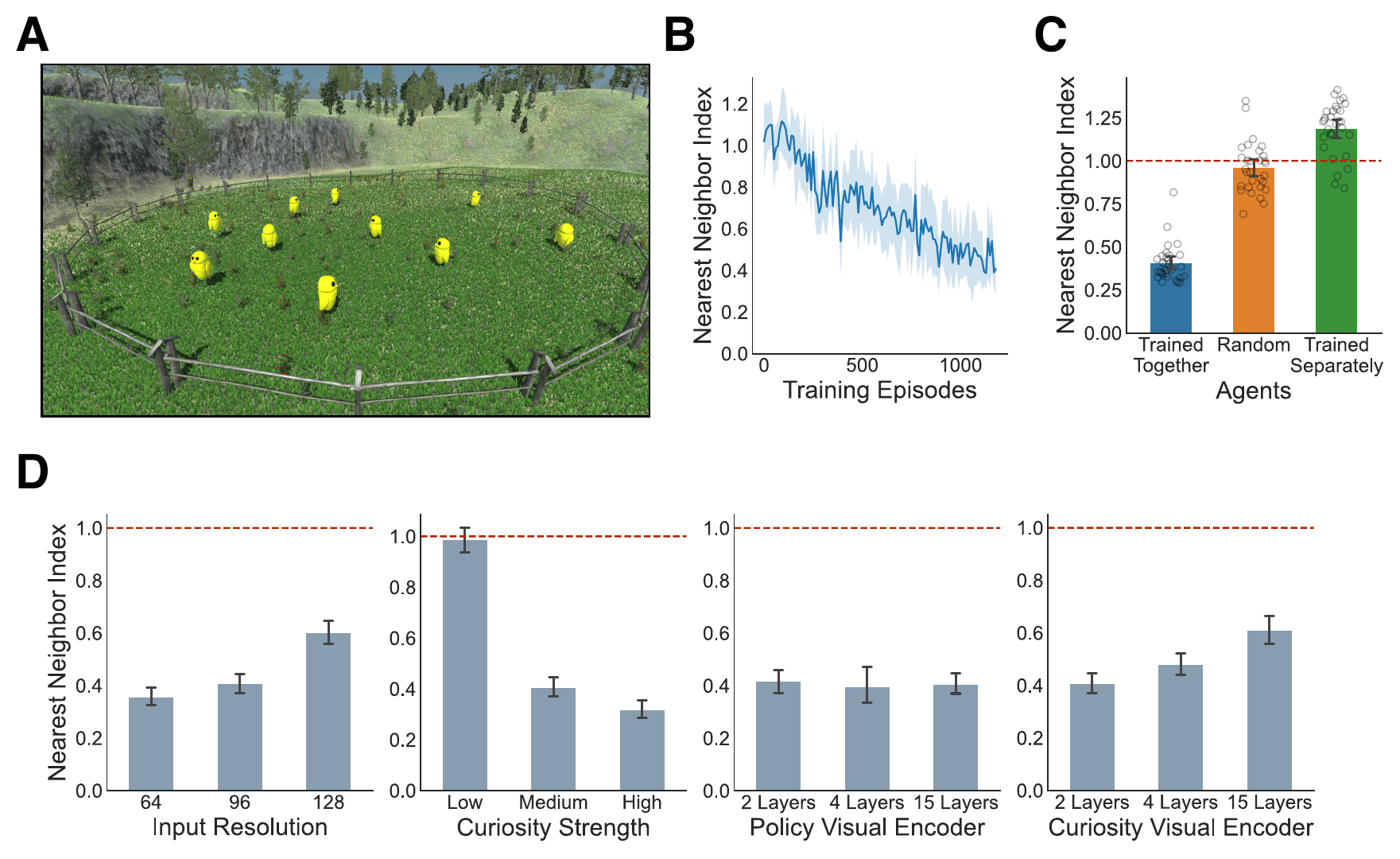}
\caption{ \textbf{Experiments 3 \& 4.} \textbf{(A)} We reared 10 artificial agents together in the same naturalistic environment. The agents were trained and tested within the fenced area. \textbf{(B)} The nearest neighbor index (NNI) during the Training Phase. As in Experiments 1-2, the agents learned to seek out other agents and exhibited grouping behavior. \textbf{(C)} Comparison of the NNIs of agents that were trained together (blue bars), agents that were trained separately (green bars), and random agents (orange bars). Only the agents that were trained together exhibited grouping behavior. \textbf{(D)} We independently varied four hyperparameters of the artificial brains: input resolution, curiosity strength, the size of the actor network encoder, and the size of the world model encoder. All of the artificial brains (except one) generated robust grouping behavior, indicating that grouping behavior can emerge in a variety of brain architectures. The one exception was the architecture with low curiosity strength. Curiosity appears to be important for producing grouping behavior. Red dashed lines represent the expected NNI of randomly distributed agents.}
\label{fig:exp34}
\end{figure*}

\subsection{Methods}
The methods were identical to those used in Experiment 1, except in the following ways. First, we trained and tested the agents in a new environment. To mimic the complexity of natural visual environments, we designed an environment referred to here as “Realistic World,” an open terrain with natural objects including grass, mountains, and trees (\cref{fig:exp34}A). Within Realistic World, we created an arena on a grassy field enclosed by a circular fence (diameter: 21m). Second, during the Training Phase, we reared one group of artificial agents (n = 10) with groupmates and another group of artificial agents (n = 10) without groupmates. If our artificial agents learned grouping behavior from their visual experiences with groupmates, then only the agents reared with groupmates should have exhibited grouping behavior in the Test Phase (when all of the agents were tested together in the same environment). We trained the agents for 1,200 episodes and tested the agents for 30 episodes.

\subsection{Results and Discussion}
The agents that were reared with groupmates developed robust grouping behavior, exhibiting collective action during the Test Phase (\cref{fig:exp34}C; NNI: M = 0.40, SD = 0.11). These agents had much lower NNIs than the random agents (mean difference = 0.56, SD = 0.13; \textit{t}(58) = 17.22, $P$ < $10^{-22}$, Cohen’s \textit{d} = 4.52). In contrast, the agents that were reared without groupmates did not exhibit collective action during the Test Phase. These agents had NNIs similar to random agents (M = 1.19, SD = 0.15) and were much more dispersed than the agents that were trained with groupmates (independent samples \textit{t}-test, \textit{t}(58) = 22.90, $P$ < $10^{-28}$, Cohen’s \textit{d} = 6.01). Thus, these artificial agents did not always exhibit collective behavior. They only did so when reared (trained) in environments with groupmates. When reared in environments without groupmates, the agents largely ignored the other agents during the Test Phase.     

This result confirms that our agents developed grouping behavior by being in proximity to their groupmates during the learning period. More generally, Experiment 3 provides additional evidence that grouping behavior can develop from two learning mechanisms: deep reinforcement learning and curiosity-driven learning. Our autonomous artificial agents developed grouping behavior from high-dimensional visual inputs in naturalistic environments, in the absence of external rewards, training labels, and explicit, hardcoded interaction rules.

\section{Experiment 4}
Experiments 1-3 indicate that at least one artificial brain architecture equipped with deep reinforcement learning and curiosity-driven learning can develop grouping behavior. We next examined whether grouping behavior can develop in a variety of artificial brains, including brains with different neural network architectures, brains with different strengths of intrinsic reward, and brains receiving different retinal input resolutions. 

\subsection{Methods}
The methods were identical those used in Experiment 3, except that we varied the types of artificial brains in the artificial agents (\cref{fig:S2}). We varied four hyperparameters in the artificial brains: (i) the architecture of the actor network’s feature encoder (2-layer, 4-layer, or 15-layer neural network), (ii) the architecture of the world model’s feature encoder (2-layer, 4-layer, or 15-layer neural network), (iii) the strength of the intrinsic reward (.01, 0.1, or 1.0), and (iv) the resolution of the retinal input (64×64, 96×96, or 128×128 pixel resolution). By varying these hyperparameters, we could explore whether grouping behavior develops in small, medium, and large neural networks and study how different components of the model shape behavior. All of the agents were trained and tested in the realistic environment from Experiment 3, since this environment most closely resembled the environments encountered by animals in nature.

\subsection{Results and Discussion}
The results are shown in \cref{fig:exp34}D. Two notable findings emerged. First, the agents developed grouping behavior with all but one of the artificial brain architectures (independent samples \textit{t}-tests versus random agents, $P$s < $10^{-12}$). Even small neural networks were sufficient to produce grouping behavior, potentially explaining why animals with relatively small brains (e.g., fish, insects) still develop grouping behavior. Second, the only artificial brain architecture that did not develop grouping behavior was the one with the lowest curiosity reward strength. Agents receiving the lowest curiosity reward strength (0.01) showed no evidence of grouping behavior (independent samples t-tests versus random agents, $t$(58) = 0.45, $P$ = .65, Cohen’s $d$ = 0.12), whereas the agents receiving moderate (0.1) and strong (1.0) curiosity reward strength showed robust grouping behavior ($P$s < $10^{-22}$). Thus, our autonomous artificial agents required moderate to strong curiosity rewards to develop collective behavior. This implies that curiosity plays a key role in the development of collective behavior.

\section{Experiment 5}
Experiments 1-4 demonstrate that autonomous artificial agents can spontaneously develop collective behavior when they are raised in groups. In the final experiment, we tested whether these agents can also develop “imprinting” behavior, an early emerging form of collective behavior seen in precocial animals. During imprinting, newborn animals learn the visual characteristics of an object simply by being exposed to the object and will subsequently recognize and selectively approach that object \cite{mccabe_visual_2019}. Imprinting helps animals quickly learn to identify and recognize caregivers and conspecifics, which can lead animals to form cohesive groups.  

\begin{figure}[t]
\centering
\includegraphics[width=\linewidth]{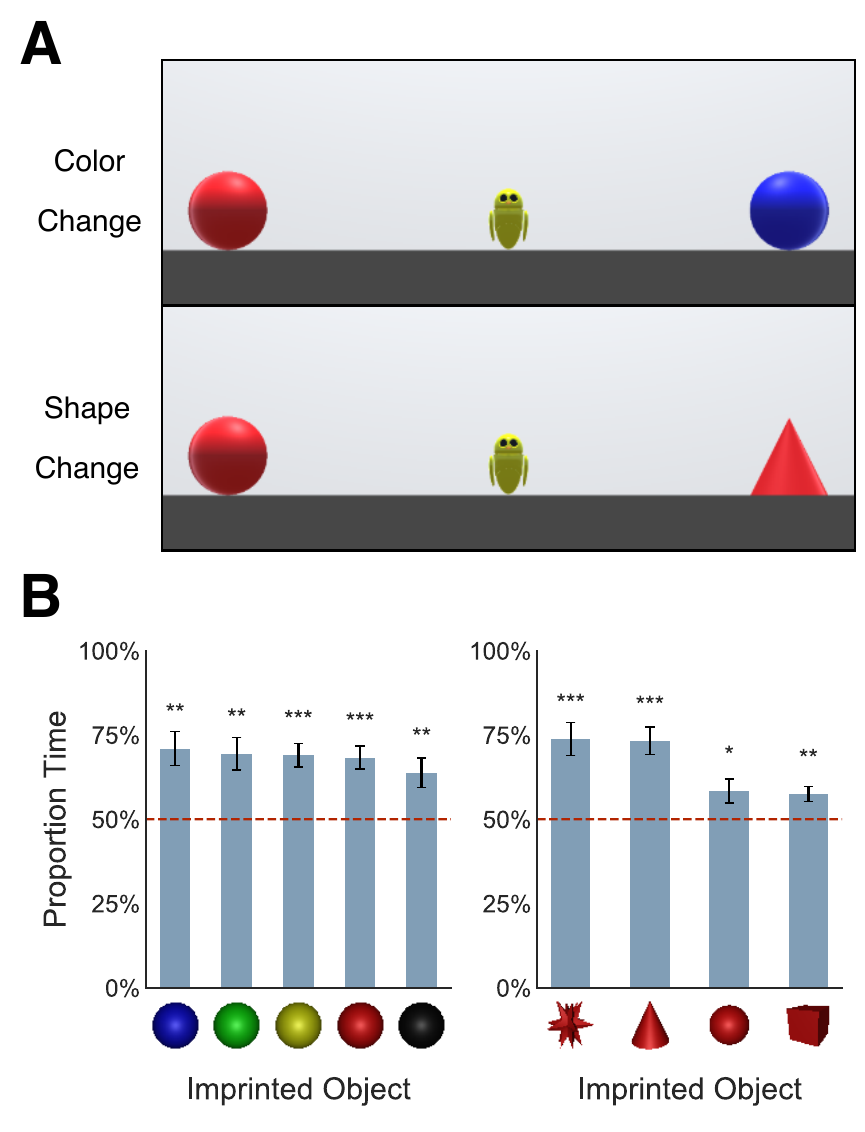}
\caption{
\textbf{Experiment 5.}
\textbf{(A)} To test the artificial agents’ color (\textit{top}) and shape (\textit{bottom}) recognition abilities, we used a two-alternative forced choice task. During the Training Phase, the agent was reared with a single “imprinted object,” which moved continuously around the chamber. During the Test Phase, the agent started each trial in the middle of the chamber at a random orientation and could approach one of two objects placed on opposite sides of the chamber. We measured whether the agents spent more time with the imprinted object than the novel object.
\textbf{(B)} Performance for each imprinted object in the color change trials (\textit{left}) and shape change trials (\textit{right}). The agents successfully distinguished their imprinted object from novel objects, using both color and shape cues. The red dashed line represents chance level (50\%). Asterisks denote statistical significance: $***P < .001, **P < .01, *P < .05$.
}
\label{fig:exp5}
\end{figure}

\subsection{Methods}
Imprinting is typically tested in the laboratory by rearing a newborn animal in a simple visual environment containing a single object, and then using a two-alternative forced-choice procedure to test whether the animal can distinguish their imprinted object from novel objects that differ along some dimension (e.g., color, shape). To mimic this experimental design, we reared artificial agents singly in a simple visual environment that contained an imprinted object. The imprinted object moved continuously around the chamber, selecting an action at random from its action space and performing that action for 10 time steps, before selecting a new random action to perform for the next 10 time steps. During the Test Phase, we used a two-alternative forced-choice procedure to test whether the agent developed a preference for the imprinted object over novel objects. On each test trial, the agent started in the center of the chamber and two objects were placed on opposite sides of the chamber. One object was the imprinted object, and the other object was identical to the imprinted object except for a change in color or shape (\cref{fig:exp5}A). The objects rotated around a vertical axis with a constant angular speed. For this experiment, we used a smaller version of Simple World with 30×30×30 unit walls. We used the same artificial brain architecture that we used in Experiments 1-3.

We measured the proportion of time the agents spent with the imprinted object versus the novel object on each trial. Each agent received 35 test trials for each test contrast (i.e., each novel color in the Color Change condition and each novel shape in the Shape Change condition). Performance was measured as mean accuracy (percent of time spent by the imprinted object versus the novel object) across the test trials.

\subsection{Results and Discussion}
The agents performed significantly higher than chance level (50\%) for all imprinted objects in both the Color Change (Fig. 5B, left; one-sample \textit{t}-tests, all $P$s < 0.01) and Shape Change conditions (Fig. 5B, right; one-sample \textit{t}-tests, all $P$s < 0.05). Thus, the agents learned to recognize their imprinted object based on both color and shape features simply by being exposed to the object during training. From a machine learning perspective, this was a challenging task because the agent only received “positive” examples (the imprinted object) during training and never received explicit labels for learning to distinguish the imprinted object from novel objects. More generally, Experiment 5 demonstrates that our model can produce the classic imprinting effect, using the same experimental design used to study imprinting in laboratory studies of newborn animals. 

This finding suggests, perhaps counterintuitively, that curiosity (i.e., a preference for what is unpredictable) can drive imprinting and grouping behavior (i.e., a preference for familiar groupmates). Note, however, that curiosity-driven deep reinforcement learning must be subject to a learning window to produce imprinting behavior; if learning stays “turned on,” the agents will continue to seek unpredictable experiences. Importantly, there is ample evidence for this type of learning window in nature. For example, filial imprinting occurs during a short sensitive period immediately after birth, when animals develop a lasting attachment to other animals seen early in life \cite{hess_imprinting_1964, horn_pathways_2004}. By learning to detect groupmates early in life, imprinting immediately provides the survival benefits associated with group living.

\section{General Discussion}
How do animals learn to group? Here we explored the possibility that grouping behavior is an emergent property of two generic learning mechanisms: deep reinforcement learning and curiosity-driven learning. To test this possibility, we built artificial brains with these two learning mechanisms, embodied those brains in virtual animal bodies, and raised those agents in realistic virtual environments. We found that these autonomous artificial agents spontaneously learned to solve the suite of tasks required for collective behavior, including ego-motion, object recognition, and a preference to approach groupmates. Like animals, our agents learned grouping behavior from raw high-dimensional visual inputs in naturalistic environments. Our agents also learned to group without external rewards or supervision, using intrinsic motivation (curiosity) to drive learning. Finally, we found that collective behavior emerges in a variety of artificial brains, including brains with small, medium, and large actor networks and brains with small, medium, and large world model networks. These results provide three main contributions to the literature.
	
 First, we show that two core learning mechanisms—deep reinforcement learning and curiosity-driven learning—are sufficient for learning how to detect, recognize, and navigate towards groupmates. Like biological organisms, our artificial agents rapidly learned these core abilities from unsupervised visual experience. Our agents self-organized their learning and spontaneously developed collective behavior. Even sparse environments, like the “Simple World” used in Experiments 1-2, provide sufficient training data for artificial agents to learn ego-motion, object recognition, and a preference for groupmates. Thus, deep reinforcement learning and curiosity-driven learning provide the computational foundations to develop collective behavior without external rewards.

Second, our results indicate that collective behavior can develop from generic learning mechanisms. By optimizing thousands of synaptic weights over thousands of visual observations, our agents developed the core skills needed for collective behavior, despite none of these abilities being explicitly encoded into the agents. Our results therefore provide a computational demonstration that it is not necessary to hardwire collective action rules into brains to produce collective behavior. When autonomous agents are equipped with deep reinforcement learning and curiosity-driven learning, and those agents are raised in realistic visual environments, collective behavior spontaneously emerges as brains “fit” themselves to the environment \cite{haber_learning_2018, hasson_direct_2020, richards_deep_2019}.

These generic learning mechanisms are inspired by the predictive processing framework \cite{clark_whatever_2013, friston_theory_2005, koster_making_2020}. According to this framework, newborn animals learn by building a predictive model of the world and iteratively updating the model to resolve uncertainties. Likewise, our artificial agents learned through prediction-based mechanisms that iteratively adjusted connection weights to resolve uncertainties. Intrinsic motivation (e.g., curiosity) also plays an important role in predictive processing \cite{koster_making_2020}. By encouraging exploration and information-seeking behavior, curiosity-driven learning allows animals to develop core behavioral skills in environments that provide little to no external rewards. Intrinsic motivation benefits developing animals, not because it is useful for solving any one particular task, but because it allows animals to scaffold their learning and build a strong foundation of knowledge that can be used to solve a wide variety of problems that the animal may encounter in the future \cite{charlesworth_intrinsically_2019}. In our study, curiosity-driven learning motivated the artificial agents to resolve errors in their internal models of the environment, allowing core visual behaviors to spontaneously develop through self-supervised learning.

Third, this study provides a pixels-to-actions model of collective behavior. Pixels-to-actions models formalize the mechanisms that underlie the entire learning process, from sensory inputs to behavioral outputs. As a result, these models can be compared against real animals, falsified, and refined. Accordingly, we see our model as a starting point towards a developmental and mechanistic understanding of collective behavior. Indeed, as we show in Experiment 4, collective behavior can emerge in a variety of model architectures within this model class. In the future, behavioral data from different animals could provide benchmarks for refining pixels-to-actions models for different species and distinguishing between competing models within this model class. Moreover, since these artificial agents develop collective behavior spontaneously, future studies could directly compare the development of collective behavior across newborn animals and artificial agents that have been raised in the same environments (and thus, learned from the same set of training data). 

\subsection{Simple versus complex models of collective behavior}
We used deep neural networks to model the development of collective behavior. However, a common criticism against deep neural networks as scientific models is that they are not directly interpretable \cite{cichy_deep_2019}. Deep neural networks are often referred to as “black box” models whose internal workings are poorly understood. In most cases, the number of parameters in a deep neural network is prohibitively large for the modeler to understand the exact contribution of each parameter to individual-level behavior. Conversely, rule-based models of collective behavior are compact and transparent. They have a handful of relevant parameters a priori linked to interaction rules which individuals are assumed to follow. It is therefore immediately apparent how changes in the model parameters affect the behavior of the agents. 

Moreover, collective animal groups are by themselves complex systems. Replacing simple rule-based models with complex deep neural networks may seem to exacerbate the “complexity problem” by introducing an additional layer of complexity without explanatory gain. Why then should we embrace deep neural networks as models for understanding collective behavior? We argue that there are four reasons why complex, pixels-to-actions models should complement (not replace) rule-based models of collective behavior.  

First, pixels-to-actions agents can serve as “task-performing” models for studying collective behavior. For decades, researchers across computational neuroscience \cite{kriegeskorte_cognitive_2018}, cognitive science \cite{newell_you_1973}, psychology 
\cite{dupoux_cognitive_2018}, and artificial intelligence \cite{hassabis_neuroscience-inspired_2017} have argued that task-performing models are essential for gaining a mechanistic understanding of information processing in brains. Recently, for example, task-performing models have been invaluable for developing predictive, executable, and neurally mechanistic models of the primate visual system \cite{schrimpf_integrative_2020}. Moreover, task-performing models focus on gaining a functional understanding of brain processes; for instance, what learning mechanisms are needed to develop collective behavior? By modeling the entire learning process in “newborn” artificial agents, researchers can test whether a candidate set of learning mechanisms are actually sufficient to solve a particular task. 

Indeed, sensory processing systems are complex, so it may be impossible to capture the role of sensory processing (e.g., visual perception) in collective behavior with simple and compact models. Real-world perception operates over high-dimensional sensory inputs, so if we want to build models that can actually perform the same tasks as animals, then we will likely need complex models. 

Second, pixels-to-actions agents can serve as integrative models, meaning that the same model can be used to study a range of behaviors. This integration is important: many historians view the integration of diverse phenomena into unified quantitative models as a hallmark of mature science. In Experiment 5, we provide one example of how a pixels-to-actions model used for one task (simulating collective behavior in groups of animals) can be used for another task (simulating the development of imprinting). Since pixels-to-actions agents are scalable (that is, they can take any retina image/video as input), we can test how these agents behave in a wide range of environments, including those not initially considered when the model was created.  

Third, it is misleading to refer to deep neural networks as “black boxes.” Rather, deep neural networks are fully transparent “glass boxes” \cite{hasson_direct_2020}. Researchers build artificial brains according to explicit architectural specifications; they train artificial brains using explicit learning rules and objective functions; they have direct access to each weight in the network; and they have full access to the training data (virtual environment) used to train the model. Given this unprecedented level of transparency, pixels-to-actions models open new experimental avenues for understanding how complex, neurally mechanistic systems learn and behave in naturalistic environments. For example, using deep neural networks, it is possible to perform techniques like deep image synthesis, in which models are used to synthesize novel “controller” images to selectively activate visual concepts in artificial brains \cite{bashivan_neural_2019}. These synthesized images can then be presented to groups of biological agents (e.g., shoals of fish) to test the models’ ability to predict and control the subjects’ behavior. If the models are accurate, then we should be able to synthesize novel controller images that change animals’ behavior in predictable ways. This approach creates a “closed-loop” experimental system between animals and computational models. This is valuable because closed-loop experimental systems can produce rapid improvements in model accuracy through a virtuous cycle of model engineering and model testing.

Finally, while deep neural networks are more complex than rule-based models, they nevertheless provide a concise framework for understanding the “neural code” \cite{hasson_direct_2020, richards_deep_2019}. Specifically, pixels-to-actions agents can be understood in terms of four well-specified components: network architecture, learning rules, objective functions, and the morphology of the agent. By specifying these four model components, we can formalize hypotheses about how sensory processing and motivational systems influence the behavior of agents, without relying on ad hoc interaction rules or simplistic sensory inputs. 

To date, it is largely unknown how changes in brain architecture, learning rules, and objective functions influence group dynamics. Future research might systematically vary these components to explore their impact on collective behavior. Likewise, changes in morphology may influence how collective behavior emerges in groups of individuals. Our approach could easily be extended to model agents with a variety of body structures (e.g. different sensory modalities, fields of view, and action spaces). It would also be interesting to explore whether pixels-to-actions models can account for other signatures of collective behavior. While our current study focused on group formation and cohesion, future studies could explore whether this model can account for other aspects of collective behavior (e.g., short-range repulsive interactions and collision avoidance, co-alignment, collective decision making, and collective intelligence) commonly observed in animal groups \cite{couzin_collective_2009}. 

On a more fundamental level, pixels-to-actions models and rule-based models of collective behavior share the principle of emergence in complex systems. Using pixels-to-actions models, we can explore how self-organizing principles of microscopic systems (system of neurons) relate to self-organization of macroscopic systems (system of individuals). Thus, we believe that pixels-to-actions models provide an essential research tool to facilitate interdisciplinary exchange across the many fields focused on collective behavior. 

In sum, we present a pixels-to-actions model of collective behavior, which indicates that we have isolated a set of learning mechanisms that are sufficient for developing collective behavior in autonomous artificial agents. Two generic learning mechanisms—deep reinforcement learning and curiosity-driven learning—can generate the core skills needed to support collective behavior in realistic visual environments. Our results complement a growing body of work using deep neural networks to model the visual \cite{yamins_performance-optimized_2014}, auditory \cite{kell_task-optimized_2018}, and motor \cite{pandarinath_inferring_2018} systems. Our results extend this approach to the study of collective action: a behavior with deep historical roots in biology, ethology, physics, and psychology.

\newpage
\begin{acknowledgements}
Funded by NSF CAREER Grant BCS-1351892 and a James S. McDonnell Foundation Understanding Human Cognition Scholar Award. We thank Brian W. Wood for help designing the artificial agents and virtual worlds, and Linda Smith and Zoran Tiganj for helpful comments on the manuscript.
\end{acknowledgements}

\section{References}
\bibliography{Grouping}

\onecolumn
\newpage
\captionsetup*{format=largeformat}
\renewcommand{\thefigure}{S\arabic{figure}}
\setcounter{figure}{0}

\section{Supplementary Information} 

\begin{figure}[hbt!]
\centering
\includegraphics[width=0.8\linewidth]{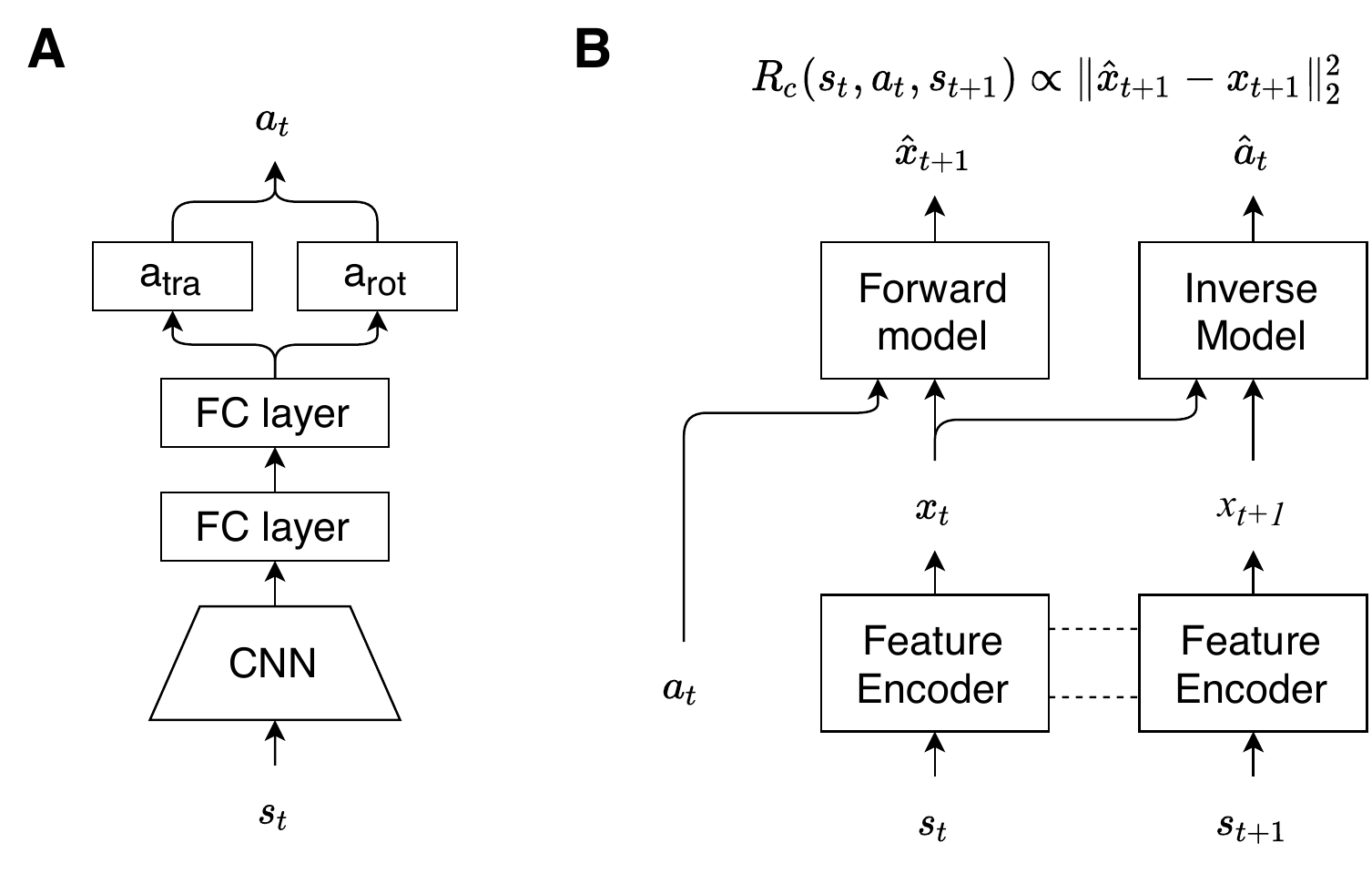}
\caption{
\textbf{Model components.}
\textbf{(A)} The actor network consisted of a convolutional neural network followed by two fully connected layers and two output heads for translation and rotation actions. 
\textbf{(B)} The world model consisted of three components: a feature encoder, an inverse model, and a forward model. The feature encoder mapped visual inputs onto a feature space. The forward and inverse models operated on the feature space to predict forward and backward dynamics. The intrinsic curiosity rewards were proportional to the prediction errors of the forward model. 
}
\label{fig:S1}
\end{figure}

\begin{figure}[hbt!]
\centering
\includegraphics[width=0.8\linewidth]{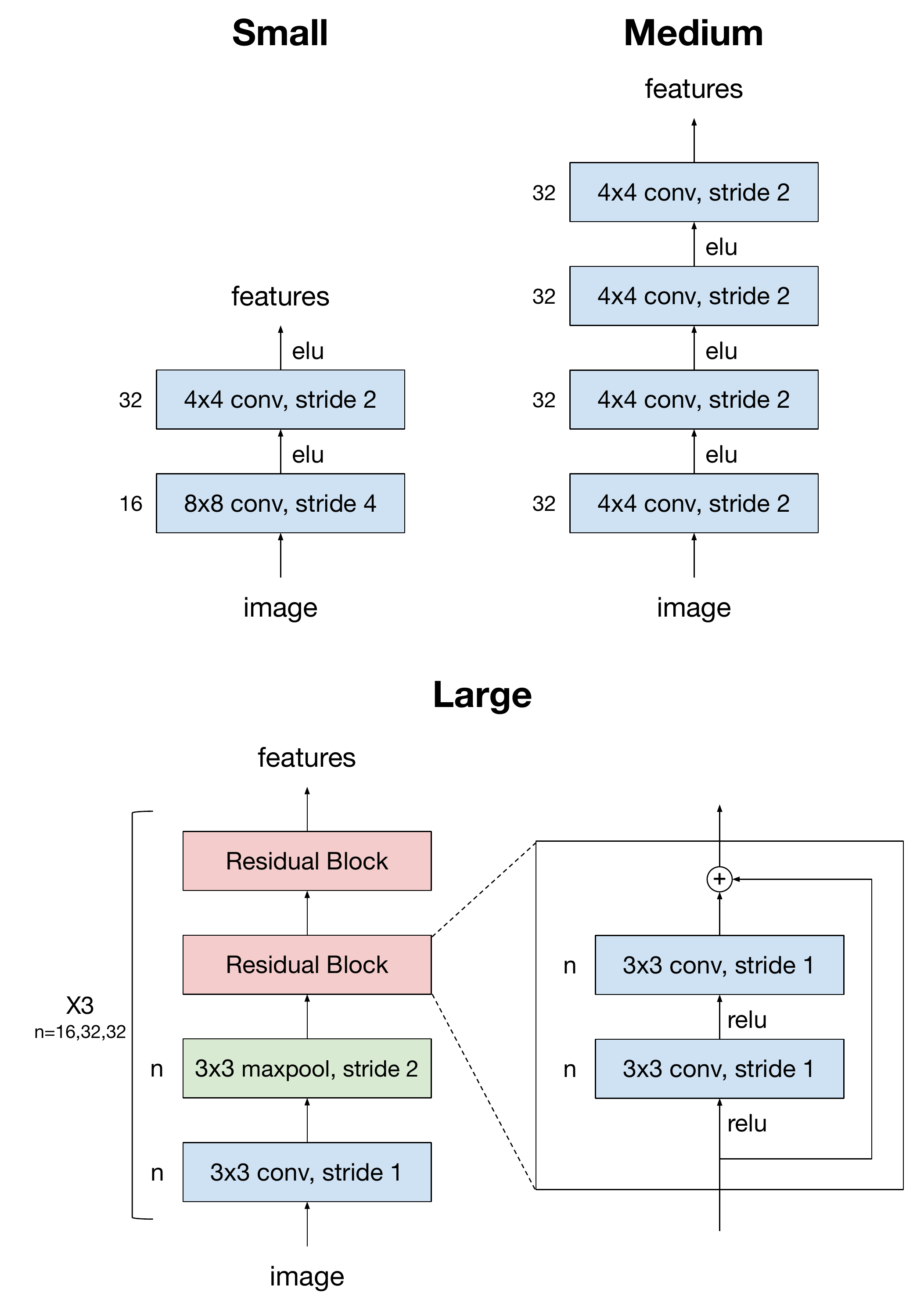}
\caption{\textbf{Visual encoder architectures.} We used three different architectures of visual encoders in Experiment 4. The Small and Medium visual encoders were 2-layer and 4-layer convolutional neural networks with ELU activations. The Large visual encoder was a 15-layer convolutional neural network with residual connections and ReLU activations.
}
\label{fig:S2}
\end{figure}

\end{document}